\pdfoutput=1

\documentclass[11pt]{article}

\usepackage[final]{acl}

\usepackage{times}
\usepackage{latexsym}

\usepackage[T1]{fontenc}

\usepackage[utf8]{inputenc}

\usepackage{microtype}

\usepackage{inconsolata}

\usepackage{graphicx}

\usepackage{hyperref}      
\usepackage{url}            
\usepackage{booktabs}       
\usepackage{amsfonts}       
\usepackage{nicefrac}       
\usepackage{microtype}     
\usepackage{xcolor}         
\usepackage{multicol}
\usepackage{multirow}
\usepackage{amsmath}
\usepackage{bbm}
\usepackage{cleveref}
\usepackage{wrapfig}
\usepackage{placeins}
\usepackage{subcaption}
\usepackage{caption}
\usepackage{array}
\usepackage{amssymb}
\usepackage{mathrsfs}
\usepackage{algorithm}
\usepackage{algorithmic}
\usepackage{xspace}

\def\AdaMoE{\raisebox{0.05ex}{\scalebox{0.96}{$\mathbf{\mathcal{A}}$}}\textit{da}\raisebox{0.04ex}{\textsc{\scalebox{0.85}{MoE}}}\xspace}

\title{\AdaMoE: Token-Adaptive Routing with Null Experts for Mixture-of-Experts Language Models}

\author{
    Zihao Zeng$^{1}$\thanks{Equal contribution.},
    Yibo Miao$^{1}$\footnotemark[1],
    Hongcheng Gao$^{2}$,
    Hao Zhang$^{3}$, 
    Zhijie Deng$^{1}$\thanks{Corresponding authors.} \\
    \textsuperscript{1}Qing Yuan Research Institute, SEIEE, Shanghai Jiao Tong University \\
    \textsuperscript{2}University of Chinese Academy of Sciences \quad \textsuperscript{3}University of California, San Diego\\
    \{zengzihao,\;miaoyibo,\;zhijied\}@sjtu.edu.cn\\
    gaohongcheng23@mails.ucas.ac.cn,\;haozhang@ucsd.edu\\
}

\begin{document}
\maketitle

\begin{abstract}
Mixture of experts (MoE) has become the standard for constructing production-level large language models (LLMs) due to its promise to boost model capacity without causing significant overheads. 
Nevertheless, existing MoE methods usually enforce a constant top-$k$ routing for all tokens, which is arguably restrictive because various tokens (e.g., ``<EOS>'' vs. ``apple'') may require various numbers of experts for feature abstraction. 
Lifting such a constraint can help make the most of limited resources and unleash the potential of the model for downstream tasks.
In this sense, we introduce \AdaMoE to realize token-adaptive routing for MoE, where different tokens are permitted to select a various number of experts. %
\AdaMoE makes minimal modifications to the vanilla MoE with top-$k$ routing---it simply introduces a fixed number of \textit{null experts}, which do not consume any FLOPs, to the expert set and increases the value of $k$. 
\AdaMoE does not force each token to occupy a fixed number of null experts but ensures the average usage of the null experts with a load-balancing loss, leading to an adaptive number of null/true experts used by each token. 
\AdaMoE exhibits a strong resemblance to MoEs with expert choice routing while allowing for trivial auto-regressive modeling. 
\AdaMoE is easy to implement and can be effectively applied to pre-trained (MoE-)LLMs.
Extensive studies show that \AdaMoE can reduce average expert load (FLOPs) while achieving superior performance. For example, on the ARC-C dataset, applying our method to fine-tuning Mixtral-8x7B can reduce FLOPs by 14.5\% while increasing accuracy by 1.69\%.
Code is available at \href{https://github.com/CengZihao/AdaMoE}{this link}.

\end{abstract}

\begin{figure*}[t]
    \centering
    \includegraphics[width=\textwidth]{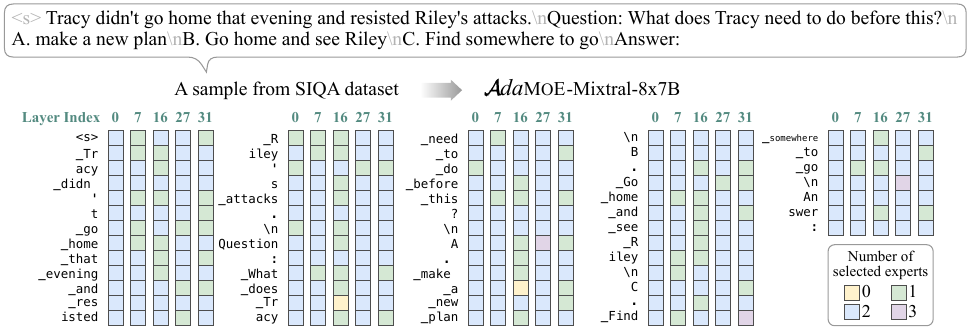}
    \caption{
    The number of selected experts for various tokens in an \AdaMoE variant of Mixtral-8x7b. 
    As shown, after applying \AdaMoE, the model possesses the ability to perform token-adaptive routing. 
    Also note that some tokens only require 1 expert for feature abstraction, which offers the opportunity for inference acceleration. 
    }
    \label{fig:visual}
\end{figure*}

\section{Introduction}
\label{sec:intro}

Large language models (LLMs) have exhibited exceptional performance across diverse tasks and domains~\citep{touvron2023llama, vicuna2023,chowdhery2023palm,zhang2022opt}. Nevertheless, LLMs’ efficacy is heavily impacted by the substantial number of parameters they possess, with some high-performing LLMs containing up to 540B parameters~\citep{chowdhery2023palm}. The mixture of experts (MoE) mechanism~\citep{shazeer2017outrageously} offers a compelling way to enhance model capability without a corresponding increase in computational overhead. Recent research further underscores the merits of MoE, vividly demonstrating its potential to support production-level applications~\citep{jiang2024mixtral,qwen_moe}.

MoE operates on the core assumption that a (small) subset of experts is sufficient to handle a single token effectively. MoE-LLMs, with Mixtral~\citep{jiang2024mixtral} and DeepSeekMoE~\citep{dai2024deepseekmoe} as popular examples, often replace the feed-forward network (FFN) in the model with a set of FFN experts. A token-level router is introduced to sparsely activate the experts for various tokens, so the computational cost is constrained to a low level. 
We can also build experts with parameter-efficient fine-tuning (PEFT) modules~\citep{hu2021lora,liu2022few} like LoRA, giving rise to Mo-LoRA approaches~\citep{zadouri2023pushing}. %

MoE routinely routes each token to a fixed amount of experts, typically the $k$ ones with top routing probabilities. %
However, not all tokens require the same number of experts for feature abstraction. 
Intuitively, the semantic tokens deserve a higher concentration of experts, while others with less significant meaning can be processed more swiftly. 
Lifting the top-$k$ routing constraint can help make the most of limited resources and unleash the potential of the model. 
To achieve this, MoE with expert choice routing~\citep{zhou2022mixture} performs expert-level routing, where each expert chooses a fixed number of tokens for processing and different tokens could be processed by different numbers of experts. %
Yet, an unacceptable drawback is that it is not suited to casual language modeling due to the reliance on future tokens for the top-$k$ token selection~\citep{zhou2022mixture}. %

This work introduces \emph{\AdaMoE}, a novel method designed to achieve token-level adaptive routing in MoE, allowing different tokens to select varying numbers of experts.
An illustrative example is presented in \Cref{fig:visual}.
\AdaMoE requires minimal changes to the vanilla MoE with top-$k$ routing by incorporating a fixed number of \emph{null experts} into the expert set. These null experts do not consume any computational resources. By increasing the value of $k$, more experts can be activated.
To encourage the average usage rate of the null experts, \AdaMoE minimizes a load balancing loss. This leads to an adaptive number of null experts and true experts being employed by each token.
Notably, \AdaMoE shares similarities with existing MoE with expert choice routing while also enabling straightforward causal language modeling.

\AdaMoE is easy to implement and can be applied to both pre-trained regular LLMs and MoE-LLMs for supervised fine-tuning. 
For the former,  we experiment on Llama2-7B~\citep{touvron2023llama} by introducing LoRA experts and corresponding routers to the model.
For the latter, we experiment on Mixtral-8x7B~\citep{jiang2024mixtral} by augmenting the original router with extra weights for the null experts.
The results underscore the effectiveness of \AdaMoE's token-adaptive mechanism in enhancing both computational efficiency and model performance. For example, when fine-tuning Llama2-7B, \AdaMoE achieves much higher accuracy across almost all evaluated datasets. Moreover, when fine-tuning Mixtral-8x7B with \AdaMoE on ARC-Challenge~\citep{allenai:arc}, we observed a 14.5\% reduction in total FLOPs, accompanied by a 1.69\% increase in accuracy.

\begin{figure*}[t]
    \centering
    \includegraphics[width=0.94\textwidth]{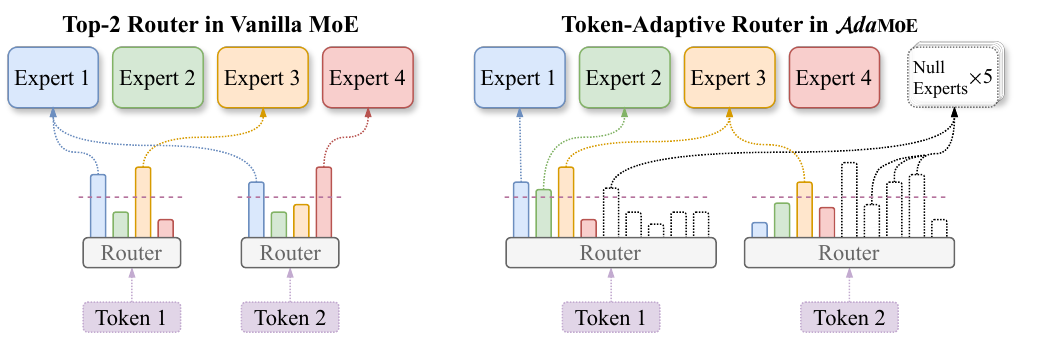}
    \caption{Comparison of Routing Mechanisms: vanilla MoE v.s. \AdaMoE.
    \textbf{Left:} In vanilla MoE, each token selects the top 2 experts based on the routing probabilities.
    \textbf{Right:} \AdaMoE introduces an additional set of \textit{null experts} and makes each token select the top 4 experts, which can include both the true and null experts. 
    For example, token 1 selects three true experts, while token 2 selects only one true expert.
    Despite this variation, the average number of true experts selected per token remains two, maintaining parity with the vanilla method.}
    \label{fig:amoe}
\end{figure*}

\section{Related Works}
\label{sec:related}

\subsection{Mixture of Experts}

Mixture of Experts (MoE)~\citep{jacobs1991adaptive, shazeer2017outrageously} is an efficient scaling technique that allows for larger model sizes with less computation, resulting in enhanced performance.
MoE models can be trained and used for inference more efficiently compared to dense models, requiring substantially fewer computational resources. 
Due to these advantages, pioneering works~\citep{jiang2024mixtral, dai2024deepseekmoe} have applied MoE to transformer-based language models and demonstrated their superiority.
Typically, they replace the feed-forward network (FFN) in each layer of the model with a routing function and multiple FFNs, referred to as experts, with only a subset of these experts being activated at any time. We refer to these models, which combine MoE and large language models (LLMs), as MoE-LLMs.

In addition to MoE-LLMs, fine-tuning techniques have also seen significant advancements.
Pre-trained LLMs are often fine-tuned for downstream tasks. However, as models increase in size, full fine-tuning becomes increasingly computationally expensive~\citep{brown2020language,chang2024survey}.
LoRA~\citep{hu2021lora} addresses this challenge by providing an effective fine-tuning methodology for scenarios with constrained computational resources.
LoRA freezes model weights and injects trainable rank decomposition matrices, thereby modifying the behavior of dense linear layers without substantially changing the original model parameters~\citep{lester2021power, an2022input}. Recent studies~\citep{zadouri2023pushing,liu2023moelora,dou2023loramoe} convincingly show that integrating LoRA with MoE offers a promising approach for achieving high performance with minimal parameter updates. Methods like MixLoRA~\citep{li2024mixlora}, MoLE~\citep{wu2023mole}, and LoRAMoE~\citep{dou2023loramoe} combine MoE with LoRA by learning multiple pairs of low-rank matrices, known as LoRA experts, and use a router to compute the probabilities of each expert for the inputs.
MoLA~\citep{gao2024higher} explores the relationship between the number of LoRA experts and the depth of model layers.
For consistency and convenience, we will refer to these methods collectively as Mo-LoRA in the following text.

\subsection{Routing Strategies}

The early MoE architecture utilized gate units as the router to select experts for each token~\citep{shazeer2017outrageously, lepikhin2020gshard}.
Following the success of the Switch Transformer~\citep{fedus2022switch} in large-scale pre-training, MoE received increased attention, leading to the development of more advanced routing algorithms.
For example, BASE Layers~\citep{lewis2021base} use a linear assignment problem to maximize token-expert affinities while ensuring each expert receives an equal number of tokens.
Hash layers~\citep{roller2021hash} employ hashing techniques on input tokens to allocate different sets of weights.
A different approach, Expert-Choice Routing~\citep{zhou2022mixture}, allows experts to select their preferred tokens, achieving a more balanced expert load and better cost-effectiveness.
Furthermore, DeepMind's Mixture-of-Depths (MoD)~\citep{raposo2024mixture} introduces a router to determine the necessity of computation for each input token at each layer.

\section{Method}
\label{sec:method}

In this section, we introduce \AdaMoE, which incorporates null experts to allow for more flexible and efficient expert selection for various tokens. 
An illustrative comparison between vanilla MoE and \AdaMoE%
is presented in \Cref{fig:amoe}.

\subsection{Preliminary on MoE}
\label{sec:pre}

MoE has been widely applied in two scenarios for large language models: MoE-LLMs~\citep{jiang2024mixtral,dai2024deepseekmoe}
and Mo-LoRA~\citep{dou2023loramoe,gao2024higher,li2024mixlora}
, as briefly introduced in \Cref{sec:related}.
The core component of both is the MoE layer, which consists of $n$ specialized experts $E_i: \mathbb{R}^{d_{in}} \to \mathbb{R}^{d_{out}}, i=1,\dots,n$ and a router $G: \mathbb{R}^{d_{in}} \to \mathbb{R}^{n}$.
The experts often have the same parameterization, such as feed-forward neural networks (FFNs) in MoE-LLMs or LoRA modules in Mo-LoRA.
The router usually activates the $k$ ($k < n$) experts with the highest routing probabilities (i.e., the top-$k$ experts) and distributes input tokens to corresponding experts.

Given an input token $x\in \mathbb{R}^{d_{in}}$, the routing process works as:
\begin{equation}
    \label{eq:router}
    G(x)\in\mathbb{R}^n:=\text{Softmax}\big(\text{TopK}(x\cdot W_g, k)\big)\ \ ,
\end{equation}
where $W_g\in\mathbb{R}^{d_{in}\times n}$ is the parameter matrix of the router,
and TopK$(\cdot,k)$ retains only the top-$k$ elements, setting the rest to $-\infty$ (so that after Softmax, the corresponding routing probabilities are zero).
The output of the MoE layer is then computed as:
\begin{equation}
    \label{eq:moe}
    y\in\mathbb{R}^{d_{out}}:=\sum\limits_{i=1}^{n}G(x)_i\cdot E_i(x)\ \ .
\end{equation}

Additionally, an auxiliary loss is applied during the training stage to encourage a balanced load across experts within the same MoE layer~\citep{fedus2022switch}.
Given a batch $\mathcal{B}$ of tokens, this load balancing loss for a MoE layer is defined as:
\begin{equation}
    \label{eq:load_loss}
    \ell_{load} := \alpha \cdot n \cdot \sum\limits_{i=1}^{n} f_i \cdot P_i\ \ ,
\end{equation}
where $\alpha$ is a hyperparameter, and $f_i$ represents the fraction of tokens dispatched to expert $E_i$,
\begin{equation}
    \label{eq:fi}
    f_i=\frac{1}{|\mathcal{B}|}\sum\limits_{x\in \mathcal{B}}\mathbbm{1}\{G(x)_i\ne 0\}\ \ ,
\end{equation}
and $P_i$ denotes the average fraction of the router probability allocated for expert $E_i$, i.e.,
\begin{equation}
    \label{eq:Pi}
    P_i=\frac{1}{|\mathcal{B}|}\sum\limits_{x\in \mathcal{B}} \text{Softmax}\big(x\cdot W_g\big)_i\ \ .
\end{equation}

\subsection{Drawback of Top-\textit{k} Router}

Almost all traditional MoE methods adopt a top-$k$ routing strategy for expert selection~\citep{fedus2022switch, lepikhin2020gshard, jiang2024mixtral}.
Therefore, each token passes through exactly $k$ experts and occupies the same amount of computation.
We first question the rationality of the fixed top-$k$ routing with the following studies.

\begin{figure}[t]
\vspace{3pt}
    \centering
    \includegraphics[width=0.49\textwidth]{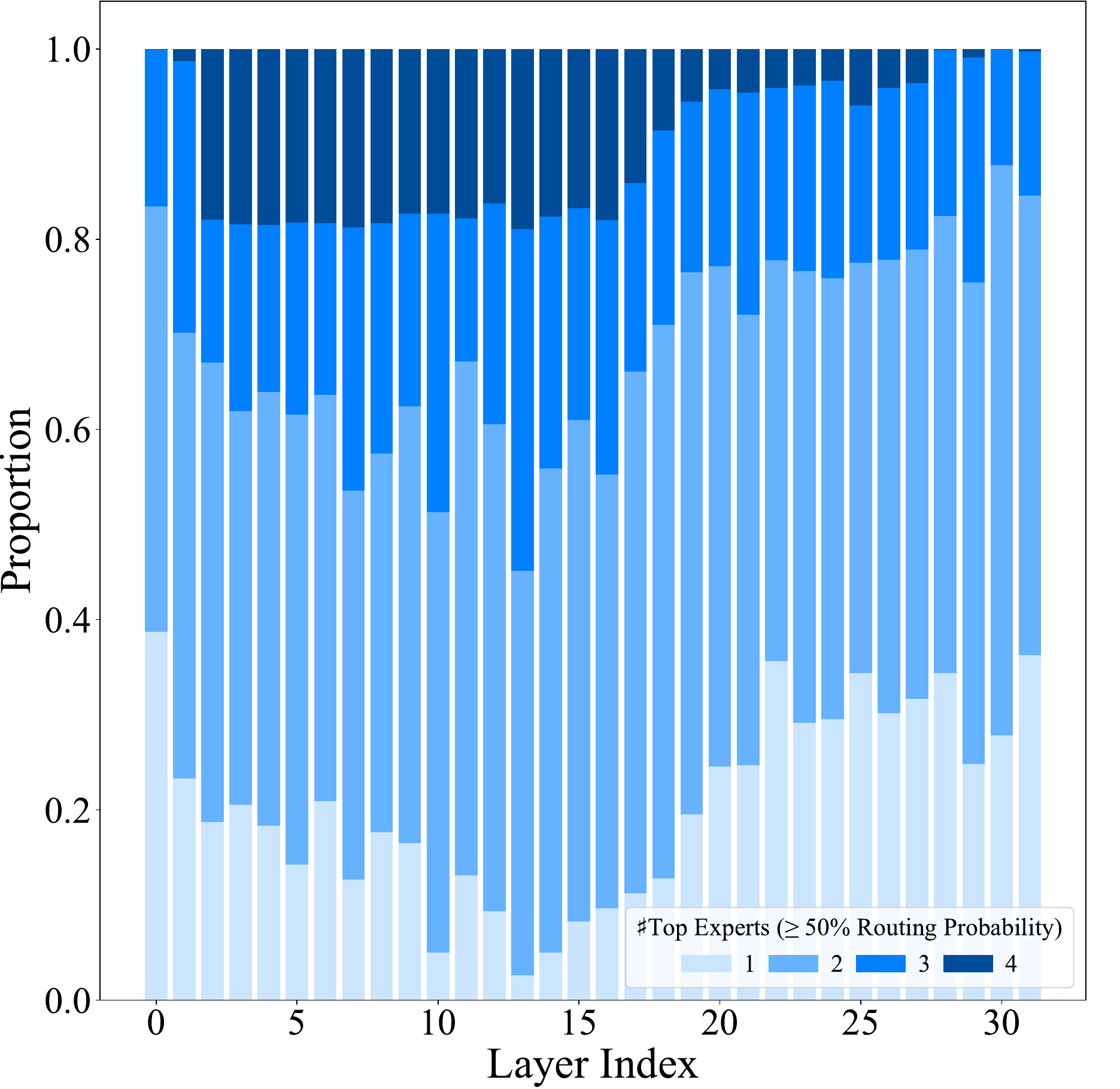}
    \caption{Proportions of the number of top experts with cumulative routing probabilities exceeding 50\% for tokens in the SocialIQA dataset.
    Each bar represents the proportion of different counts of tokens at the corresponding MoE layer in Mixtral-8x7B.
    }
    \label{fig:token}
\end{figure}

Concretely, the SocialIQA dataset~\citep{sap2019socialiqa} is fed into Mixtral-8x7B~\citep{jiang2024mixtral}, which employs a top-2 routing strategy for expert selection. 
We record the routing distribution for all tokens in each MoE layer of the model. 
To evaluate the sharpness of the routing distribution, we count the number of top experts whose cumulative routing probabilities exceed 50\% and according to this, all tokens can be divided into four categories. 
The proportions of the tokens are displayed in \Cref{fig:token}.

As shown, the proportions of tokens within different counts
show substantial variation. 
Namely, the sharpness of the routing distribution varies significantly. 
A considerable number of tokens have highly uneven routing distributions. 
Some tokens tend towards a single expert, while a significant proportion of tokens distribute attention to more than 2 experts. These observations imply that the traditional fixed top-$k$ routing strategy, which selects the same number of experts for each token, may not be optimal. This is also implied by the argument in MoD~\citep{raposo2024mixture} that some tokens may not need to pass through all MoE layers. 

\begin{figure*}[t]
    \centering
    \includegraphics[width=\textwidth]{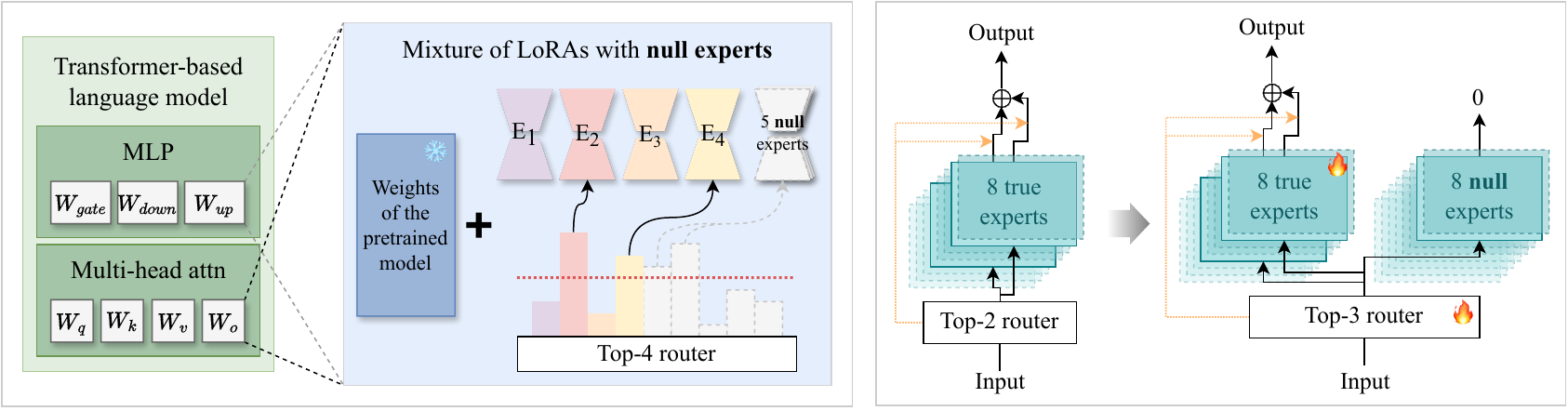}
    \caption{\textbf{Left}: Adding null experts to Mo-LoRA. \textbf{Right}: Adding null experts to the MoE layer of MoE-LLMs.%
    }
    \label{fig:compatible}
\end{figure*}

\subsection{Null Experts for Token-Adaptive Router}
\AdaMoE achieves token-adaptive expert selection by incorporating \emph{null experts}, which are defined as an empty operation requiring zero FLOPs to process the token feature. 
In the context of LLMs, common operations satisfying this requirement include a constant zero mapping and an identity mapping (we take the zero mappings null expert as the default choice in the following just for simplicity).
Consequently, an \AdaMoE layer includes $n+m$ experts, where $\{E_i\}_{i=1}^{n}$ are true experts and $\{E_i\}_{i=n+1}^{n+m}$ are null experts, and a top-$k$ router $G:\mathbb{R}^{d} \to \mathbb{R}^{n+m}$, which functions the same as the vanilla MoE router except for its output dimension.

\textbf{Token-level adaptive routing.} 
The router still performs fixed top-$k$ selection but with $k$ larger than in vanilla MoE. 
When null experts are chosen, no additional computation occurs due to their definition. 
Consequently, the number of true experts selected varies for different tokens.

\textbf{Prespecified expert load.} 
We can adjust the number of null experts according to the compute budget, and then reinforce the usage of null experts with a load balancing loss (see \Cref{sec:details}). 
This way, the load of true experts (or the overall FLOPs) can be easily adjusted to an appropriate degree.

\textbf{Autoregressive task suitability.} 
Expert-choice routing~\citep{zhou2022mixture} also allows varying numbers of experts for different tokens but struggles with autoregressive text generation since it requires considering both past and future tokens. 
In contrast, our token-choice method avoids this issue.

\textbf{Bypassing MoE layers.} 
MoD~\citep{raposo2024mixture} uses expert-choice routing to let tokens bypass some FFN layers, speeding up inference. 
Similarly, in \AdaMoE, if all selected experts for a token are null experts, 
the token effectively bypasses the \AdaMoE layer, 
achieving a similar effect.

\subsection{More Details}
\label{sec:details}

\textbf{Load balancing loss with null experts}. 
Including null experts in the load balancing loss is necessary to prevent tokens from disproportionately selecting true experts.
However, since all null experts are identical in nature, it is unnecessary to balance the load among them.
Treating null experts as distinct entities for load balancing can significantly hinder performance, as shown in \Cref{tab:modified_loss}. %

To address this, we modify the load balancing loss in \Cref{eq:load_loss} as
\begin{equation}
    \label{eq:null_loss}
    \ell_{null}=\alpha\cdot (n+m) \cdot\sum\limits_{i=1}^{n+m}  \Tilde{f}_i\cdot P_i\ \ ,
\end{equation}
where
\begin{equation*}
    \Tilde{f}_i = 
    \begin{cases} 
        f_i & \text{if } i\le n \\
        \frac{1}{m}\sum\limits_{j=n+1}^{n+m} f_j & \text{if } i > n
    \end{cases}\ \ .
\end{equation*}
By using an average load among the null experts,
we make no distinction between them, which can avoid unnecessary constraints on the router.

\textbf{Load balancing constraints: from tight to loose}.
In practice, we anneal the weight $\alpha$ of our load balancing loss to chase a better balance-efficiency trade-off. 
In particular, we first set a larger $\alpha$ to enforce strict load balancing, ensuring tokens do not disproportionately select true experts, leading to a more even load distribution among all experts.
In the latter, we use a smaller $\alpha$ to give tokens greater freedom in choosing experts. 
The empirical efficacy of doing so is verified in \Cref{tab:tight_loose}.

\textbf{Normalization of routing probabilities.}
In vanilla MoE, $\text{TopK}(x \cdot W_g, k)$ is normalized using the Softmax activation function. With null experts, we have two options: 
1) normalizing over all selected top-$k$ experts, or 
2) normalizing over only the true experts within the top-$k$ ones.
We choose the latter to ensure that the weighted average output by the \AdaMoE layer remains consistent with the scale of that from the vanilla MoE layer.

\begin{figure*}[t]
    \centering
    \includegraphics[width=\textwidth]{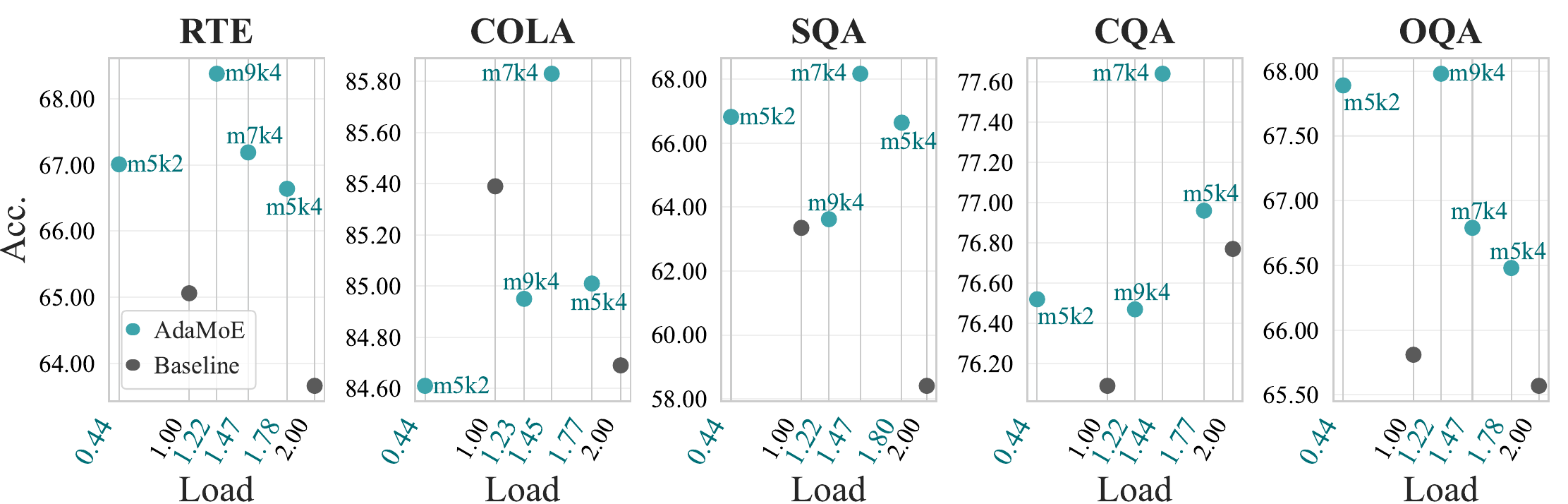}
    \caption{
    Performance comparison across five datasets: RTE, COLA, SQA, CQA, and OQA.
    The baseline is fine-tuned Llama2-7B using the vanilla Mo-LoRA method with top-1/top-2 routing. 
    Acc. represents accuracy, and Load represents the average number of experts used per Mo-LoRA module or \AdaMoE layer.
    \AdaMoE use different configurations: m5k2 (5 null experts, top-2 selection), m9k4, m7k4 and m5k4. 
    As shown, \AdaMoE achieves higher accuracy across almost all datasets compared to the baseline.
    The exact accuracy values can be found in \Cref{tab:add_exp_1}.
    }
    \label{fig:exp_1}
\end{figure*}

\subsection{Compatibility with Vanilla (MoE-)LLMs}
\label{sec:compatible}

\AdaMoE is designed to be plug-and-play, able to be seamlessly integrated with pre-trained LLMs and MoE-LLMs, as illustrated in \Cref{fig:compatible}. Due to resource constraints, we mainly focus on fine-tuning such models.
For fine-tuning regular LLMs with the Mo-LoRA architecture, we need to add a randomly initialized router and multiple LoRA experts to the corresponding module.
When applying \AdaMoE to MoE-LLMs, the router’s output dimensions are expanded to provide corresponding probabilities for null experts. 
The parameters for the new dimensions can be derived from the original parameter values.
This ensures the expanded router balances the load across all experts, including both true and null experts, at the beginning of the fine-tuning process. 
For more specific implementation details, see \Cref{sec:setup_1}. To fine-tune \AdaMoE, we need to adjust the router and experts to meet our token-adaptive routing strategy and follow the detailed modifications outlined in \Cref{sec:details} to achieve adaptive routing.

\section{Experiments}
\label{sec:exps}

In this section, we demonstrate the superior performance of \AdaMoE across various benchmarks, particularly in reducing expert load and enhancing task performance through its token-adaptive routing strategy. We first apply our method to regular LLMs with the Mo-LoRA architecture (\Cref{sec:lora_exp}). Then apply it to traditional MoE-LLMs (\Cref{sec:moe_exp}). 
Additionally, We also include extensive ablation studies to provide further insights into our approach’s effectiveness.

\subsection{Application to Regular LLMs}
\label{sec:lora_exp}

\vspace{5pt}
\subsubsection{Experiments Setup}

\textbf{Model and datasets.} 
We select Llama2-7B~\citep{touvron2023llama} as our base model due to its strong performance and popularity within the AI community. To validate the effectiveness of our method, we evaluate it on two distinct task types using five widely recognized datasets.
The first task focuses on semantic understanding, for which we use two datasets from the renowned GLUE Benchmark~\citep{wang2018glue}: Recognizing Textual Entailment (RTE) and the Corpus of Linguistic Acceptability (COLA). %
The second task involves commonsense reasoning and includes the following datasets: ScienceQA (SQA)~\citep{lu2022learn}, CommonsenseQA (CQA)~\citep{talmor2018commonsenseqa}, and OpenBookQA (OQA)~\citep{mihaylov2018can}.

\vspace{3pt}
\textbf{Baseline and implementation details.}
To highlight our method’s significance, we use the typical Mo-LoRA method as the baseline for comparison. For each MoE/\AdaMoE layer, we set $n=4$ (4 true experts). For the baseline, we set $k=1,2$ for the top-$k$ routing strategy, which are the most common choices. For our \AdaMoE, we selected various configurations for $k$ (the number of top-$k$ experts) and $m$ (the number of null experts).
We use AdamW~\citep{loshchilov2017decoupled} as the optimizer with a learning rate of 3e-4. The rank of each LoRA expert is set to 8, and the initialization of the LoRA modules follows the original LoRA implementation~\citep{hu2021lora}. For each LLM layer, we applied LoRA to $(W_q, W_k, W_v, W_o)$ in the self-attention modules and $(W_{gate}, W_{down}, W_{up})$ in the MLP modules.
We trained on each dataset for 2 epochs, using 3 random seeds, and averaged the results to obtain the final performance metrics.

\begin{table*}[t]
\setlength{\tabcolsep}{5.8pt}
  \centering
  \begin{tabular}{ccccccccc}
    \toprule
    ~ & Metric & \textbf{WINO} & \textbf{HELLA} & \textbf{PIQA} & \textbf{SIQA} & \textbf{OQA} & \textbf{ARC-C} & Avg.\\
    \midrule
  \footnotesize Original Mixtral-8x7B & \multirow{1}{*}{Acc.} & \multirow{1}{*}{55.96} & \multirow{1}{*}{53.62} & \multirow{1}{*}{68.06} & \multirow{1}{*}{64.59} & \multirow{1}{*}{65.40} & \multirow{1}{*}{83.73} & \multirow{1}{*}{65.23}\\
  \footnotesize Fine-tuned Mixtral-8x7B & \multirow{1}{*}{Acc.} & \multirow{1}{*}{80.43} & \multirow{1}{*}{84.10} & \multirow{1}{*}{\textbf{90.48}} & \multirow{1}{*}{76.36} & \multirow{1}{*}{\textbf{89.00}} & \multirow{1}{*}{87.46} & \multirow{1}{*}{84.64} \\
    \midrule
    \multirow{3}{*}{\AdaMoE} & Acc. & \textbf{81.93} & \textbf{85.50} & 90.32 & \textbf{76.97} & 88.20 & \textbf{89.15} & \textbf{85.35} \\
    ~ & \footnotesize \%FLOPs$\downarrow$ & 14.99 & 14.10 & 18.07 & 16.31 & 13.22 & 14.55 & 15.21 \\
    ~ & Load & 1.66 & 1.68 & 1.59 & 1.63 & 1.70 & 1.67 & 1.66 \\
    \bottomrule
  \end{tabular}
  \caption{
  Comparison of performance and computational efficiency across six datasets: WINO, HELLA, PIQA, SIQA, OQA and ARC-C. Metrics include Acc. (accuracy), \%FLOPs$\downarrow$ (percentage of FLOPs reduction by \AdaMoE compared to the baselines), and Load (the average number of experts used per MoE/\AdaMoE layer). The baselines are original/fine-tuned Mixtral-8x7B, both using the top-2 routing strategy (Load = 2.00). \AdaMoE not only reduces FLOPs but also achieves better accuracy across most datasets compared to the fine-tuned Mixtral-8x7B with LoRA.
  }
  \label{tab:mixtral}
\end{table*}

\vspace{-5pt}
\subsubsection{Experiments Results}
\vspace{-5pt}

The results are shown in \Cref{fig:exp_1}. We use accuracy as the main metric to evaluate the model’s performance~\footnote{LoRA expert load has minimal impact on total FLOPs; therefore, it is not considered a primary evaluation metric.}.  It is evident that \AdaMoE achieves higher accuracy across almost all datasets compared to the traditional baseline. For instance, on the RTE and OQA datasets, all configurations of \AdaMoE surpass the baseline in accuracy. This trend continues across the other datasets, demonstrating the robustness and effectiveness of \AdaMoE in achieving better performance with more adaptive expert utilization.

\subsection{Application to MoE-LLMs}
\label{sec:moe_exp}

\subsubsection{Experiments Setup}
\label{sec:setup_1}

\textbf{Model and datasets.}
We use Mixtral-8x7B~\citep{jiang2024mixtral} as the base model, where each MoE layer has 8 FFN experts and a top-2 router. We selected six well-known datasets from different categories for our experiments: WinoGrande (WINO)~\citep{sakaguchi2021winogrande} for coreference resolution, Hellaswag (HELLA)~\citep{zellers2019hellaswag}, PIQA~\citep{Bisk2020}, and SIQA~\citep{sap2019socialiqa} for commonsense reasoning, OpenBookQA (OQA)~\citep{mihaylov2018can} for reading comprehension, and ARC-Challenge (ARC-C)~\citep{allenai:arc} for science examination.

\vspace{3pt}
\textbf{Baseline and implementation details.}
Due to the substantial resources required for pre-training, we focus on fine-tuning. To save memory, we use 4-bit quantization and the QLoRA method~\citep{dettmers2024qlora}. The LoRA target modules for the baseline are \verb|gate|, \verb|w1|, \verb|w2|, and \verb|w3|.
For our \AdaMoE, we modify this architecture as described in \Cref{sec:compatible}. Specifically, we add null experts to each MoE layer, and the router expands its output dimension to assign probabilities to all experts. To simplify the modification, we define an additional module, \verb|gate2|, whose parameters can be derived from \verb|gate|.
\footnote{For instance, if \texttt{gate2} has an output dimension of 16, meaning there are 16 null experts, the parameters of \texttt{gate2} can be copied from \texttt{gate} in two segments.} 
Together, \verb|gate| and \verb|gate2| form the router that assigns weights to all experts. Thus, the LoRA target modules for our method are \verb|gate|, \verb|gate2|, \verb|w1|, \verb|w2|, and \verb|w3|.
The rank of the LoRA module is set to 8, and the learning rate is 5e-5. Due to the tendency of MoE-LLMs to overfit during fine-tuning, we use 1000 samples for training on each dataset and train for 2 epochs. In the 2 epochs, we set different values of $\alpha$ in \Cref{eq:null_loss} to $\alpha_1=0.02, \ \alpha_2=0.0001$, as described in \Cref{sec:details}.
All evaluations are conducted using OpenCompass~\citep{2023opencompass} to assess accuracy.

\begin{table*}[t]
\setlength{\tabcolsep}{7.7pt}
  \centering
  \begin{tabular}{ccccccccccc}
    \toprule
    ~ & Baseline & \multicolumn{9}{c}{\AdaMoE} \\
    \cmidrule(lr){2-2}\cmidrule(lr){3-11}
    $m,k$ & $0,2$ & $8,3$ & $16,4$ & $32,4$ & $32,5$ & $32,6$ & $40,6$ & $40,7$ & $40,8$ & $48,8$ \\
    \midrule
    Acc. & 76.36 & \textbf{76.97} & \textbf{76.92} & 66.27 & 72.93 & \textbf{76.46} & 69.86 & 76.05 & \textbf{77.23} & 74.67 \\
    Load & 2.00 & 1.63 & 1.66 & 0.77 & 1.05 & 1.54 & 1.01 & 1.49 & 1.64 & 1.48 \\
    \bottomrule
  \end{tabular}
  \caption{
  Performance of different $m$ and $k$ combinations on the SIQA dataset. The Baseline represents fine-tuned Mixtral-8x7B using LoRA method, with a Load of 2. Bold values indicate accuracy higher than the baseline.
  }
  \label{tab:siqa}
  \vspace{4pt}
\end{table*}

\subsubsection{Experiment Results}
In this section, we present the results for the configuration with $m=8$ and $k=3$ (i.e., 8 null experts and top-3 expert selection), as shown in \Cref{tab:mixtral}. Additional results are in \Cref{sec:ablation} and \Cref{tab:add_exp_2}.

\textbf{Accuracy.} \AdaMoE outperforms the baseline on WinoGrande, HellaSwag, SIQA, and ARC-Challenge. Although the baseline slightly surpasses \AdaMoE on PIQA and OpenBookQA, \AdaMoE achieves a higher average accuracy.

\textbf{FLOPs.} FFNs account for the majority of the FLOPs during inference. This issue is exacerbated in Mixtral-8x7B, which replaces the FFN with a set of 8 FFNs and selects the top-2 during each inference step. This greatly increases the computational load. \AdaMoE significantly reduces FLOPs across all datasets, achieving an average reduction of 15.21\% compared to the baseline. This demonstrates that \AdaMoE is more computationally efficient while maintaining competitive performance.

\textbf{Load.} The Load metric indicates the average number of experts used per MoE/\AdaMoE layer. The baseline method has a Load of 2. In contrast, \AdaMoE achieves a lower average Load of 1.66, indicating more efficient utilization of experts.

Overall, the results confirm the effectiveness of the token-adaptive mechanism in improving both computational efficiency and model performance.

\begin{table*}[ht]
  \centering
  \begin{minipage}[b]{0.7\linewidth}
    \centering
    \begin{tabular}{lcccccccccc}
      \toprule
      ~ & \multicolumn{2}{c}{RTE} & \multicolumn{2}{c}{COLA} & \multicolumn{2}{c}{SQA} & \multicolumn{2}{c}{OQA}\\
      \cmidrule(lr){2-3}\cmidrule(lr){4-5}\cmidrule(lr){6-7}\cmidrule(lr){8-9}
      ~ & Acc. & Load & Acc. & Load & Acc. & Load & Acc. & Load \\
      \midrule
      $\ell_{bal}$ & 56.68 & 1.77 & 83.68 & 1.77 & 65.65 & 1.78 & 69.80 & 1.76 \\
      $\ell_{null}$ & \textbf{67.51} & 1.77 & \textbf{85.01} & 1.77 & \textbf{66.64} & 1.80 & \textbf{71.40} & 1.77 \\
      \bottomrule
    \end{tabular}
    \caption{Comparison of accuracy and load on four datasets using load balancing loss with and without balancing among null experts. $\ell_{bal}$ represents the loss with load balancing constraints among null experts, and $\ell_{null}$ represents the loss without these constraints. Bold values indicate higher accuracy.}
    \label{tab:modified_loss}
  \end{minipage}%
  \hspace{0.65cm}
  \begin{minipage}[b]{0.25\linewidth}
    \centering
    \begin{tabular}{ccc}
      \toprule
      ~ & \multicolumn{2}{c}{SIQA}\\
      \cmidrule(lr){2-3}
      Option & Acc. & Load \\
      \midrule
      1) & 80.19 & 1.50 \\
      2) & 81.27 & 1.54 \\
      \bottomrule
    \end{tabular}
    \caption{
    1) Normalizing all selected top-$k$ experts, and 2) normalizing only the true experts within the top-$k$. 
    }
    \label{tab:norm}
  \end{minipage}
  \vspace{-17pt}
\end{table*}

\subsection{Ablation}
\label{sec:ablation}

In this section, we provide results for various $m$ and $k$, beyond the single configuration shown in \Cref{sec:moe_exp}. We also present ablation studies for \AdaMoE, corresponding to \Cref{sec:details}. Additional ablation experiments can be found in \Cref{sec:appendix_ablation}.

\textbf{More results for \Cref{sec:moe_exp}.}
We tested different combinations of $m$ and $k$ on the SIQA dataset, with results shown in \Cref{tab:siqa}. Compared to the $m=8,k=3$ configuration in \Cref{sec:moe_exp}, \AdaMoE can further reduce the expert load (FLOPs) while maintaining competitive performance. For example, with $m=32,k=6$, the expert load is 1.54 (79.57\% of baseline FLOPs), yet accuracy remains higher than the baseline.
There are also accuracy differences among configurations with similar loads. For instance, $m=40,k=7$ and $m=48,k=8$ have nearly identical loads but differ in accuracy. This discrepancy highlights areas for further exploration in future research.

\vspace{2pt}
\textbf{Load balancing loss with null experts.}
To verify the effectiveness of the modified load balancing loss introduced in \Cref{eq:null_loss}, we selected two datasets from each of the semantic understanding and commonsense reasoning tasks. The results, illustrated in \Cref{tab:modified_loss}, show that lifting the load balancing constraints among null experts significantly improves the performance of the fine-tuned model on the RTE, COLA, SQA, and OQA datasets.

\vspace{2pt}
\textbf{Load balancing constraints: from tight to loose.}
The effectiveness of the annealing
training process described in \Cref{sec:details} is validated in \Cref{tab:tight_loose}. The tight load balancing constraints in the first epoch effectively control the expert load in \AdaMoE, meeting our expectations. The loose constraints in the second epoch allow tokens greater  freedom 
in selecting experts, thereby enhancing performance with almost no increase in expert load. For example, on the WINO dataset, the accuracy increased by 5.69\% compared to the result after epoch 1, with almost no increase in expert load.

\begin{table}[t]
\vspace{4.5pt}
  \centering
  \begin{tabular}{cccccc}
    \toprule
     ~ & ~ & \multicolumn{2}{c}{WINO} & \multicolumn{2}{c}{SIQA} \\
    \cmidrule(lr){3-4}\cmidrule(lr){5-6}
     ~ & ~ & Acc. & Load & Acc. & Load \\
    \midrule
     \multirow{2}{*}{$\alpha_{1}$} & Baseline & 78.14 & 2.00 & 75.38 & 2.00 \\
     ~ & \AdaMoE & 76.24 & 1.65 & 75.90 & 1.62 \\
     \rule{0pt}{14pt}
     \multirow{2}{*}{$\alpha_{2}$} & Baseline & 80.43 & 2.00 & 76.36 & 2.00 \\
     ~ & \AdaMoE & 81.93 & 1.66 & 76.97 & 1.63 \\
    \bottomrule
  \end{tabular}
  \caption{Performance for finetuning Mixtral-8x7B with \AdaMoE on the WINO and SIQA datasets for two epochs with $\alpha_1 = 0.02$ and $\alpha_2 = 0.0001$.
  }
  \label{tab:tight_loose}
  \vspace{-8pt}
\end{table}

\vspace{5pt}

\textbf{Normalization of routing probabilities.} 
We tried the two options mentioned in \Cref{sec:details} on the SIQA dataset, and the results are shown in \Cref{tab:norm}. As we can see, option 2) is a superior choice, showing a significant improvement in accuracy
, with only a minor change in expert load.

\vspace{2pt}

\section{Conclusion}
\label{sec:conclusion}
\vspace{3pt}

MoE has been a promising method for training powerful models with fewer parameters. In this paper, we introduced \AdaMoE, which uses null experts to enable token-level adaptive expert allocation and overcome the drawbacks of fixed expert allocation. Extensive experiments validate its effectiveness. The \AdaMoE approach significantly enhances efficiency and adaptability, paving the way for more capable large language models.

\section*{Limitations}

One potential drawback of this work is that we did not pre-train a MoE-LLM using our \AdaMoE method. Pre-training an MoE-LLM would have allowed us to thoroughly evaluate the full capabilities and performance improvements of our method, but the significant resources required made it impractical for our current study.
Additionally, we did not explore the scenario of null experts as identity mappings, where null experts would also need zero FLOPs to process input tokens. We hypothesize that this approach might accelerate training convergence because null experts as identity mappings would potentially update their corresponding router parameters more frequently.

We acknowledge these limitations and leave these aspects for future work. Addressing these issues could provide a more comprehensive evaluation of the \AdaMoE method and potentially uncover additional benefits or areas for improvement.

\bibliography{custom}

\newpage
\appendix

\section{Additional Ablation}
\label{sec:appendix_ablation}
\vspace{-2pt}
\begin{table*}[h]
\setlength{\tabcolsep}{13pt}
  \centering
    \begin{tabular}{clrrrr}
      \toprule
      ~ & ~ & \multicolumn{2}{c}{Epoch} & \multicolumn{2}{c}{Rank of LoRAs} \\
      \cmidrule(lr){3-4}\cmidrule(lr){5-6}
      ~ & ~ & \multicolumn{1}{c}{1} & \multicolumn{1}{c}{10} & \multicolumn{1}{c}{8} & \multicolumn{1}{c}{32} \\
      \midrule
      \multirow{2}{*}{Baseline} & Acc. & 45.95 & 87.19 & 45.95 & 46.72 \\
      ~ & Load & 2.00 & 2.00 & 2.00 & 2.00 \\
      \multirow{2}{*}{\AdaMoE} & Acc. & \textbf{48.88} & \textbf{88.54} & \textbf{48.88} & \textbf{49.01} \\
      ~ & Load & \textbf{1.92} & \textbf{1.88} & \textbf{1.92} & \textbf{1.89} \\
      \bottomrule
    \end{tabular}
    \caption{Robustness of our method under different epochs and ranks of LoRAs.}%
    \label{tab:6}
\end{table*}

\textbf{Robustness.} We primarily considered the impact of two hyperparameters, the number of epochs and the rank of LoRA module. We evaluated their impact on the SQA dataset, as shown in \Cref{tab:6}.
Regardless of the number of training epochs and rank of the LoRA, our method outperforms the baseline consistently.
Therefore, we can conclude that our method demonstrates strong robustness across different epochs and ranks of LoRAs.

\vspace{-5pt}
\section{Additional Results}
\vspace{-2pt}
\begin{table*}[h]
  \centering
  \begin{tabular}{cccccccc}
    \toprule
    & & Metric & \textbf{RTE} & \textbf{COLA} & \textbf{SQA} & \textbf{CQA} & \textbf{OQA} \\
    \midrule
    \multirow{4}{*}{Baseline} & \multirow{2}{*}{k1} & Acc. & 65.06 & 85.39 & 63.35 & 76.09 & 65.81 \\
    & & Load & 1.00 & 1.00 & 1.00 & 1.00 & 1.00 \\
    \cmidrule{2-8}
    & \multirow{2}{*}{k2} & Acc. & 63.66 & 84.69 & 58.41 & 76.77 & 65.57 \\
    & & Load & 2.00 & 2.00 & 2.00 & 2.00 & 2.00 \\
    \midrule
    \multirow{10}{*}{\AdaMoE} & \multirow{2}{*}{m5k4} & Acc. & 66.64 & 85.01 & 66.64 & 76.96 & 66.48 \\
    & & Load & 1.78 & 1.77 & 1.80 & 1.77 & 1.78 \\
    \cmidrule{2-8}
    & \multirow{2}{*}{m7k4} & Acc. & 67.19 & 85.83 & 68.17 & 77.64 & 66.79 \\
    & & Load & 1.47 & 1.45 & 1.47 & 1.44 & 1.47 \\
    \cmidrule{2-8}
    & \multirow{2}{*}{m9k4} & Acc. & 68.38 & 84.95 & 63.62 & 76.47 & 67.98 \\
    & & Load & 1.22 & 1.23 & 1.22 & 1.22 & 1.22 \\
    \cmidrule{2-8}
    & \multirow{2}{*}{m5k2} & Acc. & 67.01 & 84.61 & 66.82 & 76.52 & 67.89 \\
    & & Load & 0.44 & 0.44 & 0.44 & 0.44 & 0.44 \\
    \bottomrule
  \end{tabular}
  \caption{
  Exact values for \Cref{fig:exp_1}, averaged from results with 3 random seeds.
  }
  \label{tab:add_exp_1}
\end{table*}

\textbf{Exact values for results in \Cref{fig:exp_1}.} Table \ref{tab:add_exp_1} presents the exact values corresponding to \Cref{fig:exp_1}, averaged from results with three random seeds. The performance metrics evaluated include accuracy and average number of experts used per Mo-LoRA module/\AdaMoE layer across five datasets: RTE, COLA, SQA, CQA, and OQA.

For the baseline model, two configurations (k1 and k2) are tested. The k1 configuration, with a fixed top-1 routing strategy, achieves an accuracy of 65.06 on RTE, 85.39 on COLA, 63.35 on SQA, 76.09 on CQA, and 65.81 on OQA. The k2 configuration, with a fixed top-2 routing strategy, results in slightly lower accuracy values of 63.66 on RTE, 84.69 on COLA, 58.41 on SQA, 76.77 on CQA, and 65.57 on OQA.

The \AdaMoE model is evaluated with four configurations (m5k4, m7k4, m9k4, and m5k2). The m5k4 configuration achieves an accuracy of 66.64 on RTE, 85.01 on COLA, 66.64 on SQA, 76.96 on CQA, and 66.48 on OQA, with a load ranging from 1.77 to 1.80. The m7k4 configuration shows improved accuracy, reaching 67.19 on RTE, 85.83 on COLA, 68.17 on SQA, 77.64 on CQA, and 66.79 on OQA, with a load between 1.44 and 1.47. The m9k4 configuration presents an accuracy of 68.38 on RTE, 84.95 on COLA, 63.62 on SQA, 76.47 on CQA, and 67.98 on OQA, with a load consistently around 1.22. Lastly, the m5k2 configuration records an accuracy of 67.01 on RTE, 84.61 on COLA, 66.82 on SQA, 76.52 on CQA, and 67.89 on OQA, with a significantly lower load of 0.44.

\textbf{Additional results for \Cref{sec:moe_exp}.} Table \ref{tab:add_exp_2} presents the performance of various $m$ and $k$ combinations for the \AdaMoE model across different datasets, including ARC-C, HELLA, OQA, PIQA, and WINO. These results supplement the experimental findings discussed in \Cref{sec:moe_exp}.

For the \textbf{ARC-C} dataset, the baseline model with $m=0$ and $k=2$ achieved an accuracy of 87.46 and a load of 2.00. The \AdaMoE configurations demonstrated varying performance, with the highest accuracy of 89.15 observed for $m=8$ and $k=3$, accompanied by a load of 1.67. As the values of $m$ and $k$ increased, the load generally decreased, with the lowest load of 1.34 recorded for $m=40$ and $k=8$.
On the \textbf{HELLA} dataset, the baseline model achieved an accuracy of 84.10 and a load of 2.00. The \AdaMoE model’s best performance was observed with $m=8$ and $k=3$, achieving an accuracy of 85.50 and a load of 1.68. Similar to the ARC-C results, higher values of $m$ and $k$ led to reduced loads, with a minimum load of 1.37 for $m=40$ and $k=7$.
For the \textbf{OQA} dataset, the baseline model achieved an accuracy of 89.94 and a load of 2.00. The \AdaMoE configurations showed varying results, with the highest accuracy of 89.2 observed for $m=16$ and $k=4$, and the load varying between 1.49 and 1.71. The lowest load of 1.50 was recorded for $m=40$ and $k=8$.
In the case of the \textbf{PIQA} dataset, the baseline model reached an accuracy of 90.48 and a load of 2.00. The best accuracy among the \AdaMoE configurations was 90.32 for $m=8$ and $k=3$, with a load of 1.59. The load decreased as $m$ and $k$ values increased, reaching a minimum of 1.32 for $m=40$ and $k=7$.
Finally, on the \textbf{WINO} dataset, the baseline model achieved an accuracy of 80.43 and a load of 2.00. The highest accuracy of 81.93 was observed for $m=8$ and $k=3$, with a load of 1.66. The load showed a decreasing trend with increasing values of $m$ and $k$, with the lowest load of 1.45 recorded for $m=40$ and $k=8$.

\begin{table*}[h]
\setlength{\tabcolsep}{14pt}
  \centering
  \begin{tabular}{cccccccc}
    \toprule
    \textbf{ARC-C} & Baseline & \multicolumn{5}{c}{\AdaMoE} \\
    \cmidrule(lr){1-1}\cmidrule(lr){2-2}\cmidrule(lr){3-8}
    $m,k$ & 0,2 & 8,3 & 16,4 & 24,5 & 32,6 & 40,7 & 40,8 \\
    \midrule
     Acc. & 87.46 & 89.15 & 87.12 & 86.10 & 85.08 & 86.10 & 85.76 \\
    Load & 2.00 & 1.67 & 1.70 & 1.56 & 1.49 & 1.59 & 1.34 \\
    \toprule
    \textbf{HELLA} & Baseline & \multicolumn{5}{c}{\AdaMoE} \\
    \cmidrule(lr){1-1}\cmidrule(lr){2-2}\cmidrule(lr){3-8}
    $m,k$ & 0,2 & 8,3 & 16,4 & 24,5 & 32,6 & 40,7 & 40,8 \\
    \midrule
    Acc. & 84.10 & 85.50 & 83.10 & 81.30 & 80.40 & 82.50 & 79.20 \\
    Load & 2.00 & 1.68 & 1.64 & 1.45 & 1.39 & 1.37 & 1.44 \\
    \toprule
    \textbf{OQA} & Baseline & \multicolumn{5}{c}{\AdaMoE} \\
    \cmidrule(lr){1-1}\cmidrule(lr){2-2}\cmidrule(lr){3-8}
    $m,k$ & 0,2 & 8,3 & 16,4 & 24,5 & 32,6 & 40,7 & 40,8 \\
    \midrule
    Acc. & 89 94 & 88.2 & 89.2 & 86.6 & 86.8 & 85 & 82.6 \\
    Load & 2.00 & 1.70 & 1.71 & 1.49 & 1.54 & 1.56 & 1.50 \\
    \toprule
    \textbf{PIQA} & Baseline & \multicolumn{5}{c}{\AdaMoE} \\
    \cmidrule(lr){1-1}\cmidrule(lr){2-2}\cmidrule(lr){3-8}
    $m,k$ & 0,2 & 8,3 & 16,4 & 24,5 & 32,6 & 40,7 & 40,8 \\
    \midrule
    Acc. & 90.48 & 90.32 & 89.99 & 88.30 & 86.67 & 86.78 & 85.42 \\
    Load & 2.00 & 1.59 & 1.53 & 1.46 & 1.39 & 1.32 & 1.33 \\
    \toprule
    \textbf{WINO} & Baseline & \multicolumn{5}{c}{\AdaMoE} \\
    \cmidrule(lr){1-1}\cmidrule(lr){2-2}\cmidrule(lr){3-8}
    $m,k$ & 0,2 & 8,3 & 16,4 & 24,5 & 32,6 & 40,7 & 40,8 \\
    \midrule
    Acc. & 80.43 & 81.93 & 79.32 & 78.17 & 77.66 & 71.43 & 79.16 \\
    Load & 2.00 & 1.66 & 1.72 & 1.71 & 1.73 & 1.59 & 1.45 \\
    \bottomrule
  \end{tabular}
  \caption{Performance of more $m$ and $k$ combinations on various datasets. As a supplement to the experimental results in \Cref{sec:moe_exp}.}
  \label{tab:add_exp_2}
\end{table*}

\vspace{-5pt}
\section{Additional Discussion}
\vspace{-2pt}
The top-$p$ router can also implement token-adaptive expert selection. It selects experts based on the sum of routing probabilities exceeding a threshold \( p \). This allows for a variable number of experts to be chosen for different tokens. However, compared to our \AdaMoE, this approach has the following drawbacks:
\begin{enumerate}
    \item The value of \( p \) cannot be predefined according to the compute budget, and finding an appropriate \( p \) often requires multiple attempts.
    \item It cannot enable tokens to bypass some layers.
\end{enumerate}

Moreover, our method is actually compatible with the top-$p$ approach. We can incorporate null experts and simultaneously use top-$p$. This compatibility opens up avenues for further exploration in the future.

\end{document}